\documentclass{article}

\usepackage[preprint]{neurips_2026} 

\usepackage[utf8]{inputen c} 
\usepackage[T1]{fontenc}    
\usepackage{hyperref}       
\usepackage{url}            
\usepackage{booktabs}       
\usepackage{amsfonts}       
\usepackage{nicefrac}       
\usepackage{microtype}      
\usepackage{xcolor}         
\usepackage{amsmath}        
\usepackage{graphicx}       
\usepackage{algorithm}      
\usepackage{algpseudocode}  
\usepackage{enumitem}

\title{PROWL: Prioritized Regret-Driven Optimization for World Model Learning}

\author{
  Ahmet H. Güzel\thanks{Corresponding author.} \\
  University College London AI Centre \\
  Odyssey
  \And
  Jenny Seidenschwarz \\
  Odyssey \\
  ~
  \And
  Benjamin Graham \\
  Odyssey \\
  ~
  \And
  Jonathan Sadeghi \\
  Odyssey \\
  ~
  \And
  Jeffrey Hawke \\
  Odyssey \\
  ~
  \And
  Ilija Bogunovic \\
  University of Basel \\
  University College London AI Centre
}
\begin{document}

\maketitle

\begin{abstract}
Modern action-conditioned video world models achieve strong short-horizon visual realism, yet remain unreliable on rare, interaction-critical transitions that dominate downstream planning and policy performance. Because passive demonstration data systematically under-samples these high-impact regimes, improving robustness requires actively eliciting model failures rather than relying on their natural occurrence. We introduce a KL-constrained adversarial curriculum in which a policy is trained to expose high-error trajectories of a diffusion-based world model while remaining close to the behavior distribution. The world model is continuously fine-tuned on these adversarially discovered trajectories, yielding an adversarial training loop that converts rare failures into a stable, near-distribution training signal without drifting into out-of-distribution exploitation. To maintain pressure on unresolved weaknesses as the model improves, we propose a Prioritized Adversarial Trajectory (PAT) buffer that re-ranks trajectories based on prediction error, action fidelity, and learning progress, focusing training on unresolved failure modes rather than repeatedly revisiting solved cases. We implement our approach in the MineRL framework and evaluate it on held-out out-of-distribution trajectories; PROWL improves robustness over models trained on passive data alone, reveals reward hacking behaviors under weak behavioral constraints, and demonstrates that effective adversarial world-model training critically depends on balancing exploratory failure discovery with explicit behavioral regularization. Our results suggest that scalable world models benefit not only from larger datasets, but also from selectively generating informative training data.
\end{abstract}

\section{Introduction}
\label{sec:intro}

World models aim to learn environment dynamics by predicting future observations conditioned on past states and actions. Recent diffusion-based latent dynamics models achieve impressive visual realism over long rollouts \citep{brooks2024video,genie,Hanyuan}, yet such evaluations often obscure a critical failure mode: models can produce visually plausible trajectories while failing to correctly respond to actions at short horizons. For downstream tasks such as planning and control, reliable long-horizon behavior depends on accurate local action-conditioned dynamics, and even small errors in early transitions can propagate into significant rollout divergence \citep{hafner2025trainingagentsinsidescalable,guzel2025imaginedautocurricula,alonso2024diffusionworldmodelingvisual}. Maintaining high fidelity over short horizons is therefore foundational to long-term robustness, and improving the former is the most direct path to the latter.
Improving short-horizon fidelity exposes a core limitation of passive training. Human demonstration datasets systematically under-sample rare, interaction-critical transitions, as they are dominated by common and successful behaviors. As a result, world models trained purely on passive data exhibit unpredictable errors precisely in the regimes where accuracy matters most. Addressing this limitation requires not merely more data, but the ability to actively identify and target the model's failure modes. This issue compounds as action spaces grow: in richer environments and playable interactive systems, the combinatorial coverage of any fixed dataset thins, and mechanisms that selectively generate training data for under-covered regimes become correspondingly more valuable.

In this work, we introduce \textbf{P}rioritized \textbf{R}egret-Driven \textbf{O}ptimization for \textbf{W}orld Model \textbf{L}earning (\emph{PROWL}), a constrained adversarial curriculum that transforms model errors into a structured training signal. We formulate training as a co-evolved min--max process in which an adversarial policy is optimized to expose high-error trajectories of a diffusion-based world model \citep{paired2020dennis,parkerholder2022evolving,pinto2017robust}, while a Kullback--Leibler (KL) constraint anchors the policy to a behavioral prior, preventing exploration from drifting into unrealistic out-of-distribution action sequences. The world model is continuously fine-tuned on these adversarially discovered trajectories, converting rare failure cases into stable supervision. This formulation induces a co-evolutionary dynamic between the policy and the world model. As the adversary searches for failure-inducing trajectories, it explores underrepresented regions of the action-state space --- compositional action patterns absent from both the passive dataset and the reference policy's rollouts.
In response, the world model adapts to these newly exposed patterns, expanding its coverage toward interaction-critical transitions. Rather than relying on passive data collection, PROWL actively generates targeted training data that tracks the model's current weaknesses. To sustain learning pressure as the model improves, we introduce a Prioritized Adversarial Trajectory (PAT) buffer, building on the principles of Prioritized Level Replay (PLR)~\citep{jiang2021prioritizedlevelreplay}, but adapted to prioritize trajectories based on world-model prediction error, action fidelity, and learning progress. This mechanism focuses training on unresolved failure modes while avoiding repeated exposure to already-solved cases, forming an adaptive curriculum over action-conditioned transitions.

We evaluate PROWL in the MineRL framework \citep{guss2019minerl}, using BASALT human demonstrations \citep{shah2021basalt} and a VPT-pretrained reference policy \citep{baker2022video}. Our primary held-out evaluation uses three BASALT tasks (\emph{MakeWaterfall}, \emph{BuildVillageHouse}, and \emph{CreateVillageAnimalPen}) disjoint from Phase~1 training (which used \emph{FindCave}) and from any adversarial buffer. We additionally evaluate on adversarially discovered failure cases and on strictly novel composite action modes; together these stress the failure-discovery loop directly. To test whether short-horizon improvements compound through autoregressive rollout, we also evaluate at long horizons. Throughout, we compare against a matched-compute Phase~2 fine-tuning baseline that uses the frozen VPT reference policy in place of the adversarial policy, isolating the contribution of adversarial discovery from that of additional fine-tuning compute. Our results show that PROWL improves over both passive pretraining and the matched-compute baseline across all four evaluations, that these gains transfer to longer rollouts, and that the framework reveals reward hacking~\citep{skalse2025definingcharacterizingrewardhacking} under insufficient behavioral constraints. These findings highlight that effective adversarial world-model training depends critically on balancing failure discovery with behavioral regularization.

Our contributions are summarized as follows:
\begin{itemize}[leftmargin=*,topsep=0.2em,itemsep=0.15em,parsep=0em]
    \item \textbf{Constrained adversarial curriculum for robust world-model training.} We propose a co-evolved min--max framework in which a KL-anchored policy actively discovers failure-inducing trajectories, while the world model is iteratively refined on these examples. Unlike prior world-model exploration methods that use ensemble disagreement~\citep{sekar2020planning,ball2020readypolicyone}, PROWL uses direct prediction regret against ground-truth rollouts as the exploration signal, requiring no ensemble and directly targeting where the world model is wrong.
    \item \textbf{Prioritized failure-driven curriculum.}
    We introduce a Prioritized Adversarial Trajectory (PAT) buffer that ranks trajectories using a structured scoring objective combining latent prediction error, pixel-space action fidelity, and learning progress, enabling a dynamic curriculum focused on unresolved world-model failures.
    \item \textbf{Co-evolutionary expansion of interaction regimes.}
    We show that adversarial training induces a co-evolutionary process in which the policy discovers novel, interaction-critical action compositions absent from both the passive BASALT dataset \emph{and} a matched-compute Phase~2 fine-tuning loop using the frozen VPT reference policy. This expands the effective training distribution of the world model toward hard-to-learn regimes while remaining constrained by a behavioral prior.
    \item \textbf{Empirical analysis of behavioral anchoring.} We demonstrate that improvements depend on maintaining a sufficient behavioral constraint, and characterize the trade-off between meaningful failure discovery and reward hacking \citep{skalse2025definingcharacterizingrewardhacking}.
\end{itemize}

\vspace{-0.9em}
\section{Background}
\label{sec:background}

\paragraph{Diffusion-Forcing World Models.}
Action-conditioned world models predict future observations from past states and actions. We follow latent video diffusion approaches~\citep{he2023latentvideodiffusionmodels}, operating in the latent space of a frozen VAE encoder~\citep{xing2024largemotionvideoautoencoding} and employing a diffusion transformer (DiT) backbone for sequence modeling~\citep{peebles2023scalablediffusionmodelstransformers}. Diffusion forcing~\citep{chen2024diffusionforcingnexttokenprediction} extends standard diffusion by allowing different noise levels across time steps, enabling a unified formulation of teacher-forced prediction and full-sequence generation. The training objective is applied to target frames:
\begin{equation}
    \mathcal{L}_{\mathrm{DF}}
    =
    \sum_{i}
    \left\|
    v_\theta(\mathbf{z}_i^{\sigma_i}, \sigma_i, \mathbf{c})
    -
    (\boldsymbol{\epsilon}_i - \mathbf{z}_i)
    \right\|_2^2,
\end{equation}
where the noisy latent is defined as $\mathbf{z}_i^{\sigma_i}=(1-\sigma_i)\mathbf{z}_i+\sigma_i\boldsymbol{\epsilon}_i$, with $\boldsymbol{\epsilon}_i\sim\mathcal{N}(0,I)$, and $\mathbf{c}$ denotes action conditioning. Under this interpolation, the target velocity is $\frac{d}{d\sigma_i}\mathbf{z}_i^{\sigma_i}=\boldsymbol{\epsilon}_i-\mathbf{z}_i$. We assign low noise to history frames and higher noise to target frames, enforcing causal prediction while retaining diffusion-based training stability.

\paragraph{Prioritized Replay.}
Prioritized Level Replay (PLR)~\citep{jiang2021prioritizedlevelreplay} constructs curricula by sampling instances according to both difficulty and recency:
\begin{equation}
    P(i)
    =
    (1-\rho)\,P_{\mathrm{score}}(i)
    +
    \rho\,P_{\mathrm{stale}}(i),
\end{equation}
where $P_{\mathrm{score}}$ favors high-error samples and $P_{\mathrm{stale}}$ promotes diversity by revisiting older instances. This mechanism focuses training on informative examples while avoiding overfitting to a static subset. We adopt this principle to construct a curriculum over trajectories, using world-model prediction error as the notion of difficulty.

\paragraph{KL-Regularized Policy Optimization.}
We train the adversarial policy using PPO~\citep{DBLP:journals/corr/SchulmanWDRK17}. To prevent the adversary from exploiting unrealistic actions, we regularize it toward a frozen behavioral reference policy $\pi_{\mathrm{ref}}$ using a forward-KL penalty:
\begin{equation}
    c_{\mathrm{kl}}\,
    \mathbb{E}_{s \sim d^{\pi_\phi}}
    \left[
    \mathrm{KL}\!\left(
    \pi_\phi(\cdot \mid s)
    \,\|\,
    \pi_{\mathrm{ref}}(\cdot \mid s)
    \right)
    \right],
\end{equation}
where $c_{\mathrm{kl}}$ controls the strength of the constraint. This penalty limits policy drift from the behavioral prior, helping ensure that discovered failures remain learnable rather than arising from unrealistic action sequences.

\section{PROWL Framework}
\label{sec:prowl_framework}

\begin{figure}[h]
    \centering
    \includegraphics[width=0.8\linewidth]{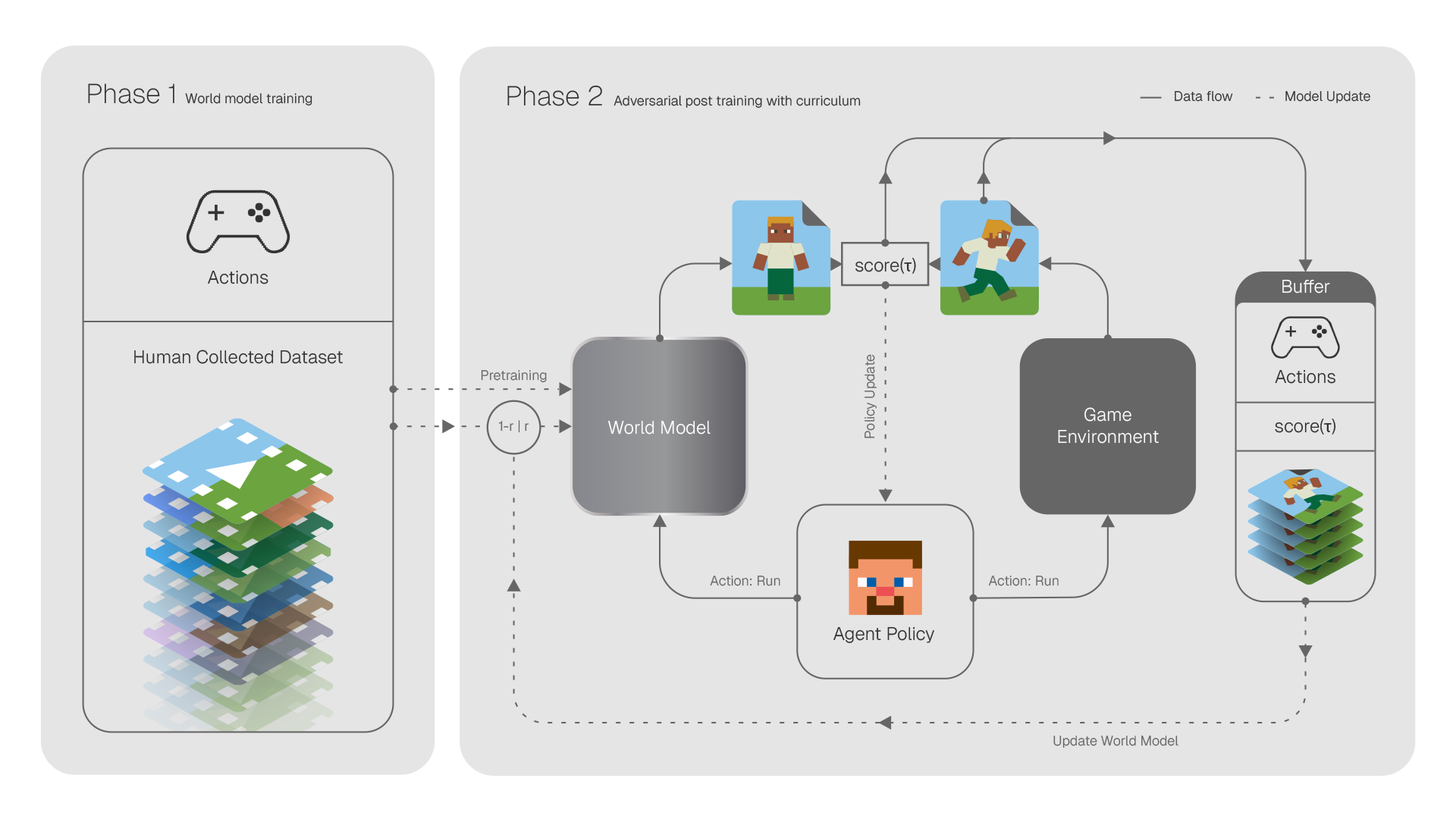}
    \caption{\textbf{PROWL framework overview.} Phase~1 pretrains the world model on the passive human-collected dataset; Phase~2 runs the adversarial curriculum, in which the agent policy acts in the game environment, the world model predicts the resulting trajectory, and per-trajectory $\mathrm{score}(\tau)$ is computed from the discrepancy between predicted and ground-truth rollouts, driving both PPO updates to the policy and insertion into the PAT buffer, which then supplies a prioritized mixture (with passive data, mixture ratio $r$) for fine-tuning the world model.}
    \label{fig:prowl_overview}
\end{figure}

PROWL is a two-phase framework that improves action-conditioned world models by actively discovering and learning from failure cases. Phase~1 trains a diffusion-transformer (DiT) world model on passive BASALT demonstrations~\citep{milani2023beddminerlbasaltevaluation}. Phase~2 introduces an adversarial curriculum in which a policy searches for trajectories that expose model failures, and the world model is iteratively refined on these trajectories via prioritized replay. Figure~\ref{fig:prowl_overview} shows the two-phase pipeline and the data/update flow between the policy, world model, and PAT buffer in Phase~2.

\paragraph{Asymmetric min--max formulation.}
PROWL can be viewed as an asymmetric two-player game between a world model $\theta$ and adversarial policy $\pi_\phi$:
\begin{equation}
\begin{aligned}
\theta^*
&=
\arg\min_\theta
(1-r)\mathbb{E}_{\tau\sim p_{\mathrm{passive}}}
[\mathcal{L}_{\mathrm{FM}}(\tau;\theta)]
+
r\mathbb{E}_{\tau\sim p_{\mathcal{B}}}
[\mathcal{L}_{\mathrm{FM}}(\tau;\theta)],
\\
\phi^*
&=
\arg\max_\phi
\mathbb{E}_{\tau\sim\pi_\phi}
[\mathrm{score}(\tau;\theta)]
-
c_{\mathrm{kl}}\,
\mathbb{E}_{s\sim d^{\pi_\phi}}
\left[
\mathrm{KL}\!\left(
\pi_\phi(\cdot\mid s)
\,\|\,
\pi_{\mathrm{ref}}(\cdot\mid s)
\right)
\right].
\end{aligned}
\label{eq:minmax}
\end{equation}

Here $p_{\mathcal{B}}$ is the sampling distribution induced by the Prioritized Adversarial Trajectory (PAT) buffer, which couples the two players through a shared memory of failure cases. This objective summarizes an alternating, memory-mediated procedure rather than a simultaneous saddle-point optimization: the policy discovers failures against a fixed model, after which the model is fine-tuned on a prioritized mixture of past failures and passive data. Algorithm~\ref{alg:prowl} summarizes this loop; the matched-compute Phase~2 (VPT-frozen) baseline used in our experiments corresponds to setting $\pi_\phi \!=\! \pi_{\mathrm{ref}}$ throughout and skipping the PPO update at line~5.
The score function $\mathrm{score}(\tau;\theta)$ aggregates per-trajectory measures of world-model prediction error, action-following fidelity, and learning progress; it is defined formally in Eq.~\ref{ref:eq_score} after the underlying error metrics (Eqs.~\ref{ref:eq_latent_regret}--\ref{ref:eq_AFS}) are introduced.

\vspace{-1.2em}
\paragraph{Chunk-level diffusion forcing.}
We train an action-conditioned DiT world model on BASALT \emph{FindCave} demonstrations using the diffusion-forcing objective (Section~\ref{sec:background}). Frames $x_{1:T}$ are encoded by a frozen VAE into latents grouped into fixed-size \emph{chunks} of $K{=}3$ latent frames, corresponding to $12$ pixel frames at $20$~fps ($0.6$~s) under the VAE's temporal compression ratio of $4$. History frames receive low noise, target frames higher noise, and the loss is applied only on the target chunk, yielding a causal chunk-level prediction objective.
\vspace{-1.2em}
\paragraph{Backbone and action conditioning.}
The backbone is the publicly released Wan2.1-T2V-1.3B diffusion transformer~\citep{wan2025wanopenadvancedlargescale}, originally a text-to-video model whose cross-attention pathway we repurpose for action conditioning: each chunk's per-frame actions are serialized into short text tokens fed through the frozen UMT5-XXL encoder~\citep{chung2023unimaxfairereffectivelanguage} (full scheme in Appendix~\ref{app:action_serialization}). Phase~1 fits all $\sim$1.3B Wan2.1 parameters with the VAE and UMT5 encoder frozen; Phase~2 (Section~\ref{sec:method_phase2}) restricts updates to the action-conditioning subset.
\vspace{-1.0em}
\subsection{Phase 2: Adversarial Curriculum}
\label{sec:method_phase2}
The adversary is initialized from the pretrained VPT-2$\times$ foundation model~\citep{baker2022video} and operates in the MineRL action space; the same checkpoint remains frozen throughout Phase~2 as the behavioral reference $\pi_{\mathrm{ref}}$. For \textbf{autoregressive rollout}, the policy generates actions $a_{0:S+H}$; the first $S$ chunks form a ground-truth seed and the remaining $H$ define the rollout. The world model predicts autoregressively at the chunk level over a fixed Wan2.1-shaped context window of $\mathrm{max}_t{=}21$ latent slots ($\le 7$ chunks), with $K{=}3$ target slots updated per denoising pass and the window sliding forward by $K$ after each chunk. Full rollout mechanics are presented in  Appendix~\ref{app:rollout}.
\vspace{-1.0em}
\paragraph{Prediction-error score.}
We compute latent regret in the VAE latent space:
\begin{equation}
\label{ref:eq_latent_regret}
\ell_{\mathrm{regret}}(\tau)
=
\sqrt{
\frac{1}{H C N_{\mathrm{lat}}}
\sum_{t=S}^{S+H-1}
\left\|
z_t^{\mathrm{pred}}-z_t^{\mathrm{real}}
\right\|_2^2
},
\end{equation}
and the Action-Follow Score (AFS) on decoded pixel frames at native MineRL resolution:
\begin{equation}
\label{ref:eq_AFS}
\ell_{\mathrm{AFS}}(\tau)
=
\frac{1}{N\cdot H_{\mathrm{raw}}}
\sum_{t=0}^{N-1}\sum_{i=1}^{H_{\mathrm{raw}}}
\left\|
F_{t,i}^{\mathrm{pred}}-F_{t,i}^{\mathrm{real}}
\right\|_2,
\end{equation}
where flows are computed using SEA-RAFT~\citep{wang2024searaftsimpleefficientaccurate}. In Eq.~\ref{ref:eq_AFS}, $H_{\mathrm{raw}} \!=\! H_{\mathrm{pix}}\!\cdot\!W_{\mathrm{pix}} \!=\! 480\!\cdot\!832$ is the per-frame pixel count (distinct from the latent-space height $H$ in Eq.~\ref{ref:eq_latent_regret}), $N \!=\! T_{\mathrm{pred}}\!-\!1$ is the number of consecutive pixel-frame pairs over which optical flow is computed (distinct from the latent-frame count $N_{\mathrm{lat}}$ in Eq.~\ref{ref:eq_latent_regret}), and $F^{\mathrm{pred}}_{t,i}, F^{\mathrm{real}}_{t,i}\!\in\!\mathbb{R}^2$ are the predicted and ground-truth flow vectors at pixel $i$ for the $t$-th consecutive frame pair. The trajectory score is:
\begin{equation}
\label{ref:eq_score}
\mathrm{score}(\tau)
=
z_{\mathcal{B}}(\ell_{\mathrm{regret}})
+
\lambda_{\mathrm{AFS}}\,z_{\mathcal{B}}(\ell_{\mathrm{AFS}})
+
\Delta\ell_{\mathrm{regret}}.
\end{equation}
Here $z_{\mathcal{B}}(\cdot)$ denotes z-score normalization over the current PAT buffer. The term $\Delta\ell_{\mathrm{regret}}$ is the signed change in the same trajectory's latent regret across consecutive buffer rescore cycles, so trajectories that have been solved lose priority while stagnant or regressing trajectories remain emphasized. The two error terms are complementary: $\ell_{\mathrm{regret}}$ penalizes global latent-plausibility errors, while $\ell_{\mathrm{AFS}}$ penalizes pixel-space motion-fidelity errors that latent agreement may not capture. The weight $\lambda_{\mathrm{AFS}}$ controls which failure mode the adversary preferentially hunts, and is one of the two axes that defines the regime spectrum analyzed in Section~\ref{sec:results_and_discussion}. We motivate the choice of optical-flow over direct pixel differences in Appendix~\ref{app:afs_justification}.
\vspace{-1.0em}
\paragraph{Policy optimization.}
We optimize $\pi_\phi$ with PPO using $\mathrm{score}(\tau)$ as a terminal reward. The loss includes a forward-KL penalty:
\begin{equation}
    \mathcal{L}_{\mathrm{PPO}}
    =
    \mathcal{L}_{\mathrm{clip}}
    +
    c_v \mathcal{L}_{\mathrm{value}}
    -
    c_e \mathcal{H}(\pi_\phi)
    +
    c_{\mathrm{kl}}
    \mathbb{E}_{s \sim d^{\pi_\phi}}
    \left[
    \mathrm{KL}\!\left(
    \pi_\phi(\cdot \mid s)
    \,\|\,
    \pi_{\mathrm{ref}}(\cdot \mid s)
    \right)
    \right].
    \label{eq:ppo_loss}
\end{equation}
The KL term constrains the adversary to remain near the frozen behavioral reference, preventing high-error trajectories from degenerating into unrealistic action sequences.
\vspace{-1.0em}
\paragraph{World-model update.} During Phase~2 fine-tuning, we freeze the Wan2.1 spatial-temporal backbone and update only the cross-attention layers and the UMT5 action-text adapter --- the action-conditioning subset, $\sim$280M of the $\sim$1.3B parameters trained in Phase~1. This targets the architectural locus of action conditioning and preserves the visual prior built during pretraining; we revisit this choice in Section~\ref{sec:results_limitations}. Updates occur every $\mathrm{T}_{\mathrm{wm}}$ iterations using a mixture of PAT-buffer and passive samples drawn per-sample with probability $r{=}0.5$, and the buffer is rescored after each update.
\vspace{-1.0em}
\paragraph{Sweep configurations.} We train six PROWL configurations along two axes: $c_{\mathrm{kl}}\in\{0.5,1.0,1.5\}$ at $\lambda_{\mathrm{AFS}}{=}0.25$ (varying the KL anchor at the default action-follow weight) and $\lambda_{\mathrm{AFS}}\in\{0.05,0.10\}$ at $c_{\mathrm{kl}}{=}1.0$ (varying the action-follow weight at the default anchor). The two productive operating points referenced throughout --- \texttt{lam010} ($c_{\mathrm{kl}}{=}1.0$, $\lambda_{\mathrm{AFS}}{=}0.10$) and \texttt{kl150} ($c_{\mathrm{kl}}{=}1.5$, $\lambda_{\mathrm{AFS}}{=}0.25$) --- share the $c_{\mathrm{kl}}{=}1.0$/$\lambda_{\mathrm{AFS}}{=}0.25$ anchor; their regime structure is analyzed in Section~\ref{sec:results_and_discussion}.

\section{Experiments}
\label{sec:results_and_discussion}

We evaluate whether PROWL improves action-conditioned prediction where passive training is weakest. We compare three world-model checkpoints: (i)~the Phase~1 pretrained model (trained on BASALT \emph{\emph{\emph{FindCave}}} only); (ii)~a matched-compute Phase~2 fine-tuning baseline that uses the same loop, optimizer, passive-data mixture ratio $r{=}0.5$, and total update budget as PROWL, but generates rollouts from the frozen VPT reference policy rather than the KL-anchored adversarial policy (denoted \emph{Phase~2 (VPT-frozen)}); and (iii)~PROWL. The Phase~2 (VPT-frozen) baseline isolates adversarial discovery from additional fine-tuning compute.
\vspace{-1.25em}
\paragraph{A regime spectrum across $(c_{\mathrm{kl}}, \lambda_{\mathrm{AFS}})$.}
The two curriculum-design parameters of PROWL define a principled axis: $c_{\mathrm{kl}}$ controls how far the adversary may drift from the behavioral reference, and $\lambda_{\mathrm{AFS}}$ biases the trajectory score toward motion-sensitive failures. Sweeping this axis (Section~\ref{sec:method_phase2}) exposes three qualitatively distinct regimes.(i)~An \emph{unanchored} regime ($c_{\mathrm{kl}}{=}0.5$) in which the adversary reward-hacks via camera-thrashing rather than discovering learnable failures (Section~\ref{sec:results_policy}). (ii)~A \emph{broad-exploration} regime, \texttt{lam010} ($c_{\mathrm{kl}}{=}1.0$, $\lambda_{\mathrm{AFS}}{=}0.10$): moderate anchor, latent-regret-weighted score; predicted to expose the widest variety of action compositions and therefore to improve where coverage matters most --- generalization to held-out tasks and strictly novel composite modes. (iii)~A \emph{focused-specialist} regime, \texttt{kl150} ($c_{\mathrm{kl}}{=}1.5$, $\lambda_{\mathrm{AFS}}{=}0.25$): tighter anchor, motion-emphasized score; predicted to concentrate on behaviorally valid hard cases, improving cross-buffer adversarial transfer, in-distribution stability, and long-horizon predictions where local-dynamics errors compound. We test each regime on the evaluations its design predicts: \texttt{lam010} on Tables~\ref{tab:wm_main} and~\ref{tab:wm_novel}, \texttt{kl150} on Table~\ref{tab:wm_kl150}. \textbf{Cross-buffer evaluation protocol.} Each checkpoint $\theta^{(a)}$ is evaluated on the off-diagonal buffers $\mathcal{B}^{(b)}, b{\neq}a$, and averaged. Off-diagonal gains thus reflect transfer to failures discovered by other adversaries rather than memorization. \textbf{Metrics and evaluation sets.} We report latent regret (Eq.~\ref{ref:eq_latent_regret}), AFS-EPE (Eq.~\ref{ref:eq_AFS}), and LPIPS; lower is better. Short-horizon evaluation uses $N_{\mathrm{seed}}{=}2$, $N_{\mathrm{pred}}{=}2$ ($2.4$\,s); long-horizon extends to $N_{\mathrm{pred}}{=}18$ ($10.8$\,s). Per-table sample sizes and dataset details appear in each caption. Since PROWL targets rare high-error regimes, we report both full-holdout means and the hardest tertile by Phase~1 latent regret.
\vspace{-1.0em}
\subsection{Adversarial Policy Dynamics}
\label{sec:results_policy}

\begin{figure}[h]
    \centering
    \includegraphics[width=1.0\linewidth]{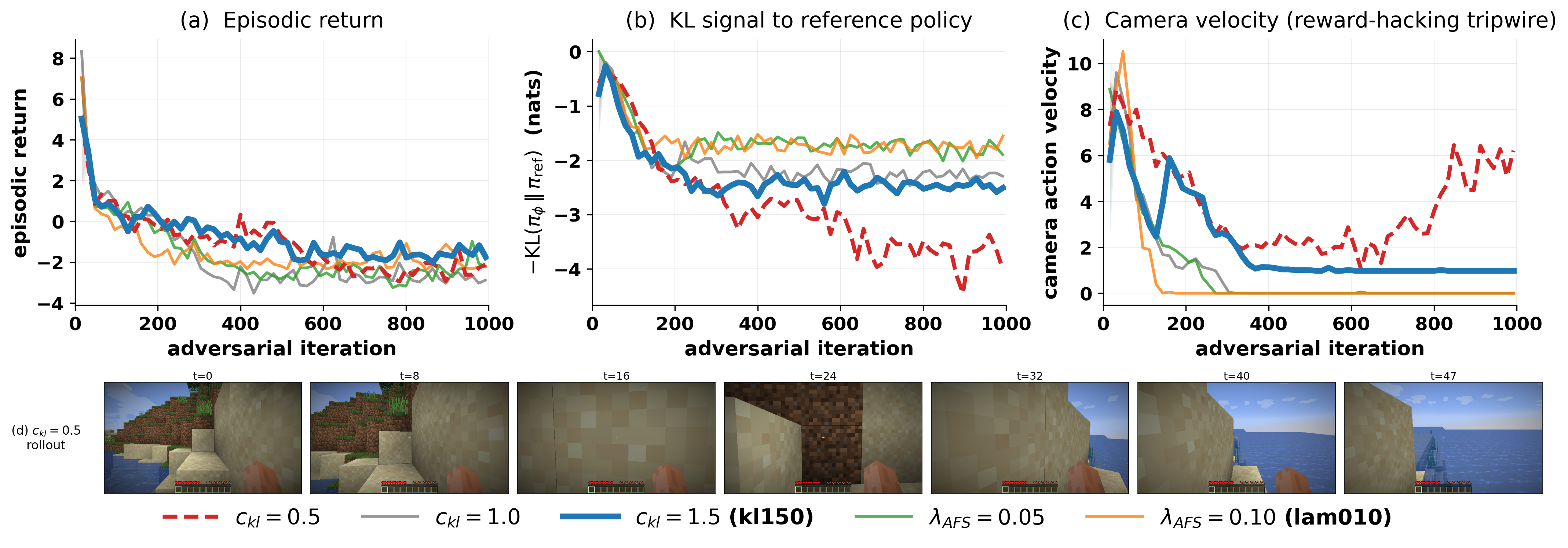}
    \caption{\textbf{Adversarial policy dynamics.} \textbf{(a)}~Episodic return (sanity check; non-stationary by construction). \textbf{(b)}~KL divergence to the frozen VPT reference separates weakly constrained exploration from stable adversarial exploration. \textbf{(c)}~Camera-action velocity, a tripwire for motion-saturation reward hacking. \textbf{(d)}~Filmstrip from the $c_{\mathrm{kl}}{=}0.5$ arm showing camera rotation reward hacking.}
    \label{fig:agent_dynamics}
\end{figure}

We ask whether the adversary discovers useful failures or collapses into reward hacking. Since the reward is the world model's prediction error, return is non-stationary by construction; we therefore interpret it jointly with two diagnostics relative to the frozen VPT reference: the Schulman-$k_3$ KL estimator~\citep{schulman2020approximating,DBLP:journals/corr/SchulmanWDRK17}, and camera action velocity (rollout-mean absolute change in the discretized camera bin), which acts as a tripwire for motion-saturation reward hacking. Figure~\ref{fig:agent_dynamics} shows the result: the weakly constrained $c_{\mathrm{kl}}{=}0.5$ arm separates from the stable arms on KL and rebounds late in training to roughly twice their camera velocity --- camera-thrashing that is technically learnable but displaces capacity from useful regimes and surfaces as visual artifacts rather than improved dynamics. Stronger KL constraints suppress this failure mode, confirming that high prediction error alone is not a useful reward signal: the adversary must expose failures the world model can productively incorporate. This matches the novelty-and-learnability criterion of~\citep{hughes2024openendednessessentialartificialsuperhuman}: KL-anchored regimes deliver both, while the unanchored regime delivers only novelty.

\vspace{-.7em}
\subsection{Zero-Shot Generalization to Held-Out BASALT Tasks}
\label{sec:results_wm}

We evaluate generalization on three BASALT tasks held out from Phase~1 training (which used \emph{FindCave} only): \emph{MakeWaterfall}, \emph{BuildVillageHouse}, and \emph{CreateVillageAnimalPen}. None is in-distribution for any of our training data sources, including the VPT-frozen baseline's rollout source. PROWL (\texttt{lam010}) improves over both Phase~1 and the matched-compute baseline on every metric, with the gap widening on the hardest subsets --- confirming the broad-exploration prediction that wider action-space coverage transfers to held-out tasks.

\begin{table}[h]
    \centering
    \caption{\textbf{Held-out generalization across three disjoint BASALT tasks.} Cross-task mean over $n{=}300$ human clips ($100$/task). The Top~25\% ($n{=}75$) and Top~10\% ($n{=}30$) subsets contain the clips with the highest Phase~1 latent regret within each task --- the hardest cases for the pretrained world model. PROWL (\texttt{lam010}) improves over both baselines on every metric, with gains widening on the harder subsets; lower is better.}
    \label{tab:wm_main}
    \small
    \setlength{\tabcolsep}{1pt}
    \renewcommand{\arraystretch}{1.15}
    \begin{tabular}{l c c c c c c c c c}
        \toprule
       & \multicolumn{3}{c}{Mean} & \multicolumn{3}{c}{Top 25\%} & \multicolumn{3}{c}{Top 10\%} \\
        \cmidrule(lr){2-4} \cmidrule(lr){5-7} \cmidrule(lr){8-10}
        Method & Lat~$\downarrow$ & AFS~$\downarrow$ & LPIPS~$\downarrow$ & Lat~$\downarrow$ & AFS~$\downarrow$ & LPIPS~$\downarrow$ & Lat~$\downarrow$ & AFS~$\downarrow$ & LPIPS~$\downarrow$ \\
        \midrule
        Phase~1 WM & 0.6514 & 28.16 & 0.5089 & 0.8804 & 32.39 & 0.6270 & 0.9888 & 37.14 & 0.6717 \\
        Phase~2 (VPT-frozen) & 0.6446 & 25.61 & 0.4990 & 0.8182 & 27.87 & 0.5935 & 0.9173 & 31.56 & 0.6414 \\
        PROWL (\texttt{lam010}) & \textbf{0.6288} & \textbf{24.60} & \textbf{0.4952} & \textbf{0.8102} & \textbf{26.07} & \textbf{0.5880} & \textbf{0.9026} & \textbf{29.38} & \textbf{0.6333} \\
        \midrule
        $\Delta$\% \texttt{lam010} vs Phase~1 & $\mathbf{-3.5\%}$ & $\mathbf{-12.6\%}$ & $\mathbf{-2.7\%}$ & $\mathbf{-8.0\%}$ & $\mathbf{-19.5\%}$ & $\mathbf{-6.2\%}$ & $\mathbf{-8.7\%}$ & $\mathbf{-20.9\%}$ & $\mathbf{-5.7\%}$ \\
        $\Delta$\% \texttt{lam010} vs VPT-frozen & $\mathbf{-2.5\%}$ & $\mathbf{-3.9\%}$ & $\mathbf{-0.8\%}$ & $\mathbf{-1.0\%}$ & $\mathbf{-6.5\%}$ & $\mathbf{-0.9\%}$ & $\mathbf{-1.6\%}$ & $\mathbf{-6.9\%}$ & $\mathbf{-1.3\%}$ \\
        \bottomrule
    \end{tabular}
\end{table}
\vspace{-1.0em}
\subsection{Focused-specialist regime: cross-buffer, stability, and long-horizon transfer}
\label{sec:results_kl150}

The focused-specialist regime, PROWL\texttt{(kl150)}, predicts gains on behaviorally valid hard cases and on long-horizon predictions where local-dynamics errors compound. This regime delivers the largest gains over the matched-compute baseline of any evaluation in this paper, with up to 8.9\% reduction in AFS-EPE on cross-buffer adversarial trajectories. Table~\ref{tab:wm_kl150} tests this prediction on three evaluations: cross-buffer adversarial trajectories drawn from all six PAT buffers , held-out \emph{FindCave} clips testing in-distribution stability, and an 18-chunk autoregressive rollout to 10.8~s. PROWL\texttt{(kl150)} improves over the matched-compute baseline on every panel, confirming all three predictions. The cross-buffer protocol rules out self-buffer memorization on the adversarial panel; the \emph{FindCave} gains confirm that the $r{=}0.5$ passive mixture prevents catastrophic forgetting; and the long-horizon panel shows that local-dynamics improvements compound through autoregressive rollout, where the matched-compute baseline alone barely improves over Phase~1.

\begin{table}[h]
    \centering
    \caption{\textbf{Focused-specialist regime: hard-case transfer, in-distribution stability, and long-horizon compounding.} Three evaluations test whether \texttt{kl150}'s tighter anchor delivers the predicted gains. \emph{Left:} cross-buffer adversarial transfer --- mean over $384$ trajectories ($64$ from each of six PAT buffers, evaluated off-diagonal so each checkpoint is scored only on buffers it did not train on). \emph{Middle:} \emph{FindCave} pretraining stability --- $n{=}64$ held-out clips from the Phase~1 training distribution. \emph{Right:} long-horizon compounding --- LPIPS and SSIM at frame $200$ of an 18-chunk autoregressive rollout ($n{=}30$ held-out BASALT clips). PROWL (\texttt{kl150}) improves over the matched-compute baseline on all three panels; lower is better except SSIM.}
    \label{tab:wm_kl150}
    \footnotesize
    \setlength{\tabcolsep}{2.5pt}
    \renewcommand{\arraystretch}{1.15}
    \begin{tabular}{l|ccc|ccc|cc}
        \toprule
        & \multicolumn{3}{c|}{Cross-buffer adv.\ ($n{=}384$)} & \multicolumn{3}{c|}{Held-out \emph{FindCave} ($n{=}64$)} & \multicolumn{2}{c}{Long-horizon ($n{=}30$)} \\
        Method & Lat$\downarrow$ & AFS$\downarrow$ & LPIPS$\downarrow$ & Lat$\downarrow$ & AFS$\downarrow$ & LPIPS$\downarrow$ & LPIPS$\downarrow$ & SSIM$\uparrow$ \\
        \midrule
        Phase~1 WM & 0.8261 & 39.87 & 0.5589 & 0.5566 & 28.96 & 0.4385 & 0.7432 & 0.4226 \\
        VPT-frozen & 0.7265 & 35.26 & 0.5484 & 0.5444 & 28.17 & 0.4298 & 0.7447 & 0.4491 \\
        PROWL \texttt{kl150} & \textbf{0.6846} & \textbf{32.13} & \textbf{0.5344} & \textbf{0.5255} & \textbf{26.09} & \textbf{0.4245} & \textbf{0.6947} & \textbf{0.4685} \\
        \midrule
        $\Delta$\% vs Phase~1 & $-17.1$ & $-19.4$ & $-4.4$ & $-5.6$ & $-9.9$ & $-3.2$ & $-6.5$ & $+10.9$ \\
        $\Delta$\% vs VPT-frozen & $\mathbf{-5.8}$ & $\mathbf{-8.9}$ & $\mathbf{-2.6}$ & $\mathbf{-3.5}$ & $\mathbf{-7.4}$ & $\mathbf{-1.2}$ & $\mathbf{-6.7}$ & $\mathbf{+4.3}$ \\
        \bottomrule
    \end{tabular}
\end{table}

\subsection{Curriculum Dynamics and Novel Interaction Modes}
\label{sec:results_curriculum}

\begin{figure}[h]
    \centering
    \includegraphics[width=0.95\linewidth]{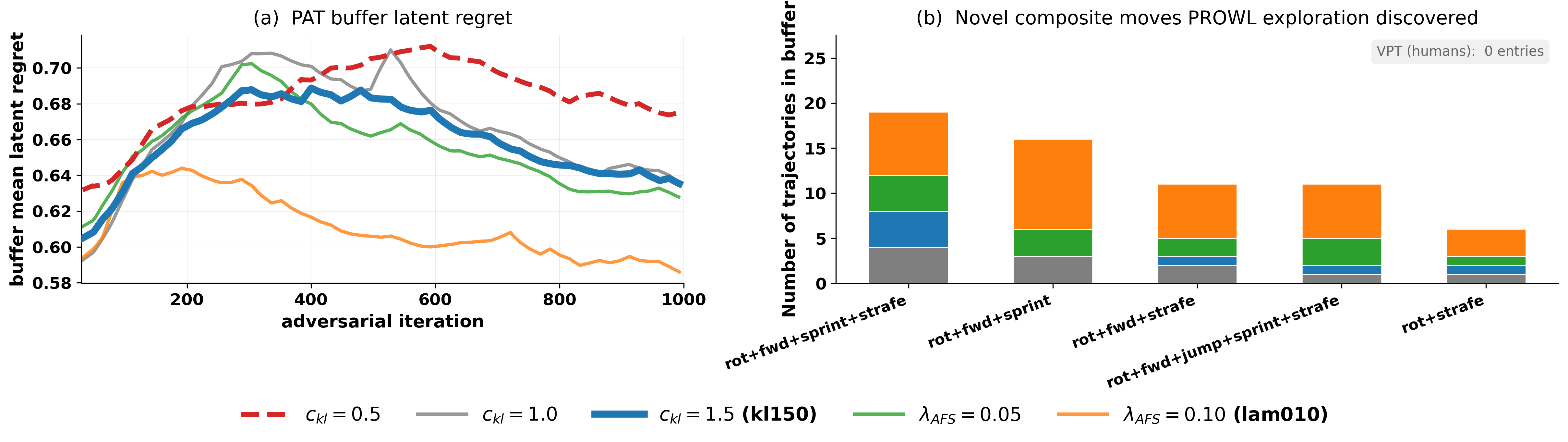}
    \caption{\textbf{Curriculum formation and novel interaction discovery.}
    \textbf{(a)}~Mean PAT-buffer latent regret. Successful arms first increase regret by discovering hard cases, then reduce it as the world model learns; the $c_{\mathrm{kl}}{=}0.5$ arm remains high because it exploits camera-thrashing shortcuts.
    \textbf{(b)}~Trajectory counts for the 5 most populated of 27 strictly novel composite modes, aggregated across all PROWL PAT buffers; the $c_{\mathrm{kl}}{=}0.5$ arm contributes zero entries (zero contribution to novel composites). }
    \label{fig:curriculum_dynamics}
\end{figure}

Figure~\ref{fig:curriculum_dynamics}(a) visualizes the min--max curriculum: buffer latent regret rises early as the adversary expands into higher-error regions, then falls as the world model incorporates the discovered trajectories. The $c_{\mathrm{kl}}{=}0.5$ arm deviates --- its regret stays high, driven by camera-thrashing that inflates error without yielding learnable dynamics --- separating useful adversarial exploration from reward hacking. We then ask whether adversarial generation produces action compositions that passive and non-adversarial fine-tuning never reach. We summarize each rollout window by a deterministic action fingerprint (full definition in Appendix~\ref{app:fingerprint}) and apply identical thresholds to BASALT, the VPT-frozen Phase~2 buffer, and all PROWL buffers. A composite mode is \emph{strictly novel} if it has zero occurrences in both the BASALT/VPT data and the VPT-frozen buffer; by construction, gains on these modes are attributable to adversarial discovery rather than additional fine-tuning compute. Across the 27 strictly novel modes (Table~\ref{tab:wm_novel}), PROWL (\texttt{lam010}) improves the cross-mode mean by $5.5\%/11.5\%/3.4\%$ over Phase~1 and $4.5\%/9.1\%/3.5\%$ over the matched-compute baseline. Figure~\ref{fig:curriculum_dynamics}(b) shows the per-arm decomposition for the most populated modes: \texttt{lam010} dominates discovery, while $c_{\mathrm{kl}}{=}1.5$ contributes less than $c_{\mathrm{kl}}{=}1.0$ --- consistent with a tighter KL anchor reducing behavioral drift, the trade-off that defines the focused-specialist regime. Together with Table~\ref{tab:wm_main}, this confirms the broad-exploration prediction: \texttt{lam010}'s moderate anchor and latent-regret-weighted score bias the adversary toward motion-sensitive failures that both expose novel composites and transfer across held-out distributions.

\begin{table}[h]
    \centering
  \caption{\textbf{Mean improvement across $27$ unique combination of strictly novel composite action modes.} Mean $\Delta$\% for PROWL (\texttt{lam010}) against Phase~1 and the matched-compute baseline; negative is better.}
    \label{tab:wm_novel}
    \small
    \setlength{\tabcolsep}{6pt}
    \renewcommand{\arraystretch}{1.15}
    \begin{tabular}{l c cc cc cc}
        \toprule
        & & \multicolumn{2}{c}{$\Delta$\% Latent} & \multicolumn{2}{c}{$\Delta$\% AFS-EPE} & \multicolumn{2}{c}{$\Delta$\% LPIPS} \\
        \cmidrule(lr){3-4} \cmidrule(lr){5-6} \cmidrule(lr){7-8}
        & $n$ & vs P1 & vs P2 & vs P1 & vs P2 & vs P1 & vs P2 \\
        \midrule
        Mean across novel modes & 27 & $\mathbf{-5.5}$ & $\mathbf{-4.5}$ & $\mathbf{-11.5}$ & $\mathbf{-9.1}$ & $\mathbf{-3.4}$ & $\mathbf{-3.5}$ \\
        \bottomrule
    \end{tabular}
\end{table}

\begin{figure}[h]
    \centering
    \vspace{0.2em}
    \includegraphics[width=0.9\linewidth]{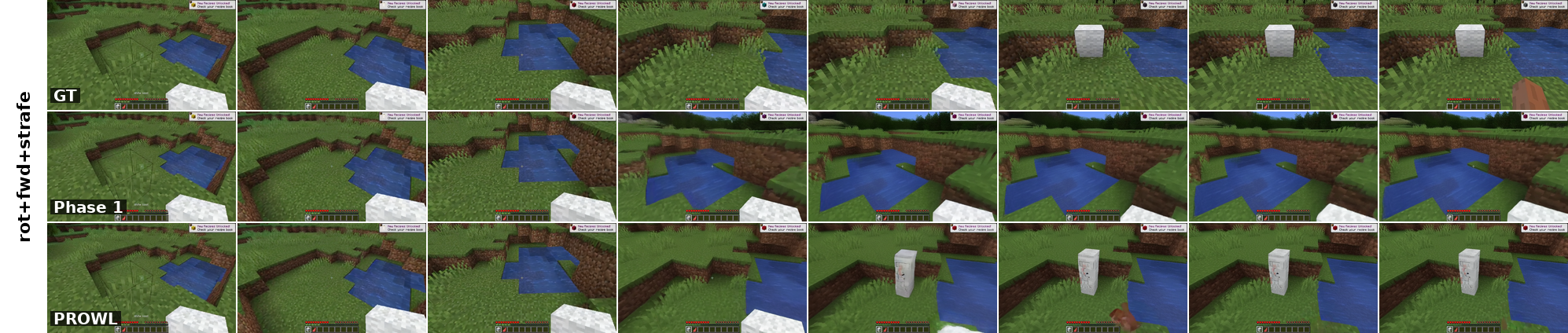}
    \vspace{0.2em}
    \includegraphics[width=0.9\linewidth]{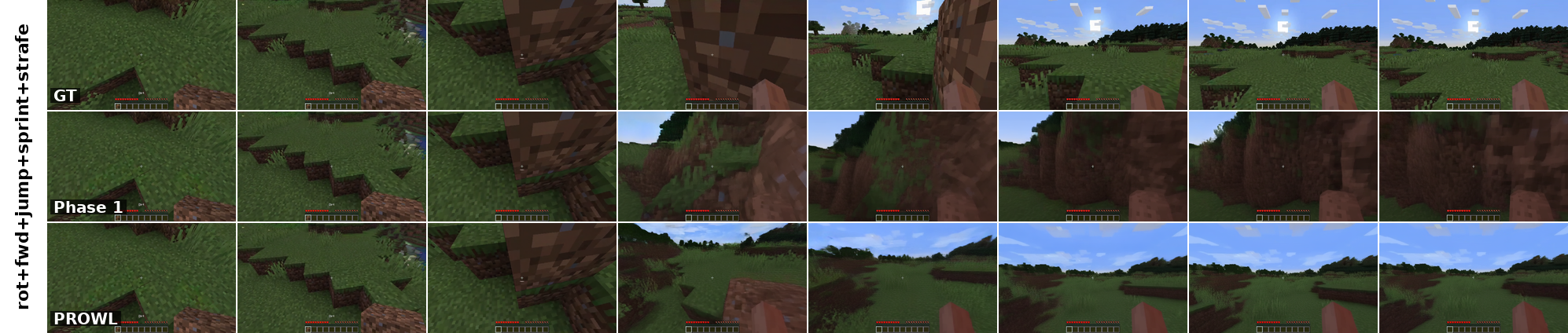}
      \caption{\textbf{Qualitative comparison on held-out and novel regimes.} Top: high-error held-out trajectory from the VPT reference policy. Remaining rows: novel composite modes discovered by PROWL, ordered by active button count. Each row shows ground truth, Phase~1, and PROWL (\texttt{lam010}) under identical actions; per-mode statistics in Table~\ref{tab:wm_novel}.}
    \label{fig:qualitative_actions}
\end{figure}

Figure~\ref{fig:qualitative_actions} shows representative rollouts under identical action sequences. The Phase~1 world model often freezes, lags behind camera motion, or predicts visually plausible but action-inconsistent transitions. PROWL better preserves the intended rotation and locomotion, supporting the quantitative finding that lower latent regret and AFS-EPE correspond to improved action following rather than only better frame appearance.

\section{Related Work}
\paragraph{World models for reinforcement learning.}
World models serve as learned simulators for planning and sample-efficient RL~\citep{oh2015action,ha2018world}, with the Dreamer family~\citep{hafner2020dream,hafner2022mastering,hafner2024mastering,hafner2025trainingagentsinsidescalable} and transformer/diffusion variants~\citep{robine2023twm,micheli2023iris,alonso2024diffusionworldmodelingvisual} training policies on imagined rollouts. PROWL inverts this relationship: the policy is not trained to solve a downstream task inside the model, but to expose the model's predictive failures so the world model itself can improve.

\paragraph{Action-conditioned video world models.}
Recent work treats world models as action-conditioned video generators for games, driving, and robotics~\citep{decart2024oasis,guo2025mineworldrealtimeopensourceinteractive,wang2023drivedreamer,gao2024vista,wu2022daydreamer,yang2023unisim}, often addressing long-horizon consistency via memory or context mechanisms~\citep{chen2024diffusionforcingnexttokenprediction,yu2025context,xiao2025worldmem,lian2025loopnav}. Our focus is complementary: we target short-horizon action-response failures directly through adversarial discovery.

\paragraph{Active data acquisition.}
Plan2Explore~\citep{sekar2020planning} and Ready Policy One~\citep{ball2020readypolicyone} actively collect data via ensemble disagreement or model variance. PROWL differs in signal, constraint, and scale: it uses realized prediction regret against ground-truth rollouts (no ensemble), KL-anchors the adversary to a behavioral prior so failures remain learnable, and operates at 1.3B parameters where exhaustive action-space coverage is infeasible.

\paragraph{Adversarial curricula and prioritized replay.}
Curriculum learning selects examples by difficulty or learning progress~\citep{bengio2009curriculum,graves2017automated}; PLR samples levels by learning potential and staleness~\citep{jiang2021prioritizedlevelreplay}; UED and adversarial RL generate challenging environments to improve agent robustness~\citep{pinto2017robust,paired2020dennis,parkerholder2022evolving,guzel2025imaginedautocurricula}. PROWL adapts these to the world model itself: the prioritized buffer trains the simulator, not the agent inhabiting it.
\vspace{3.0em}
\section{Limitations}
\label{sec:results_limitations}

\textbf{First, not all adversarial data is useful.} The unanchored regime ($c_{\mathrm{kl}}{=}0.5$) discovers high-error trajectories primarily through camera-thrashing shortcuts, yielding weak transfer despite high buffer regret --- prediction error alone is not a sufficient training signal, and the adversary must remain within behaviorally meaningful regimes for improvements to generalize. \textbf{Second, single-seed runs per configuration: a deliberate trade-off.} Given the cost of full PROWL runs, we allocated compute to mapping the $(c_{\mathrm{kl}},\lambda_{\mathrm{AFS}})$ axis rather than to point-estimate variance. Multi-seed evaluation remains future work. \textbf{Third, the long-horizon evaluation lies outside the training horizon} where PROWL applies pressure. The long-horizon panel of Table~\ref{tab:wm_kl150} should therefore be read as evidence that local-dynamics improvements compound through autoregressive rollout rather than as a direct optimization target; extending adversarial pressure to longer horizons without inducing instability remains open. Cross-environment generalization beyond MineRL/BASALT is also untested. We see this as a step toward world-model training that complements dataset scaling with selective data generation, and a natural next step is scale itself: as backbones grow and action spaces expand toward embodied and open-world settings, targeted failure discovery should become more valuable rather than less, since passive coverage thins precisely where adversarial pressure concentrates.

\vspace{-1.0em}
\section{Conclusion}
We introduce PROWL, an adversarial training framework that improves action-conditioned world models by actively discovering and replaying high-error trajectories. A KL-anchored policy exposes prediction failures, the world model is fine-tuned on them, and a prioritized buffer keeps the curriculum focused on what the model has not yet learned. Across four evaluations --- held-out BASALT tasks (Table~\ref{tab:wm_main}), cross-buffer adversarial failures, pretraining stability, and long-horizon prediction (all in Table~\ref{tab:wm_kl150}), and strictly novel composites (Table~\ref{tab:wm_novel}) --- PROWL improves over both passive pretraining and a matched-compute non-adversarial baseline. The matched-compute comparison is what makes the contribution legible: the gap between the two Phase~2 checkpoints is what adversarial discovery adds beyond additional fine-tuning compute. As action spaces grow combinatorially toward playable game systems and embodied agents, static datasets cover an ever smaller slice. Our results suggest that adversarial discovery is one mechanism for selectively generating the data that matters --- provided it is properly anchored: weakly constrained adversaries surface trajectories the model cannot productively learn from, while KL-anchored adversaries expose failures that are both hard and behaviorally valid.

\bibliographystyle{plainnat}
\bibliography{references}

\newpage
\appendix
\section{Implementation Details and Hyperparameters}
\label{app:hyperparams}

Table~\ref{tab:hyperparams} reports the hyperparameters used across all PROWL runs.
All sweep arms share these values; only $c_{\mathrm{kl}}$ and $\lambda_{\mathrm{AFS}}$
differ across arms, as described in the sweep configuration paragraph in
Section~\ref{sec:method_phase2}. The Phase~1 initialization checkpoint was selected
by scanning checkpoints at 2k, 4k, 6k, 8k, and 10k steps using latent regret, AFS-EPE,
and LPIPS; we used the 10k-step checkpoint, which achieved the best LPIPS.

\begin{table}[!t]
    \centering
    \caption{\textbf{PROWL hyperparameters.} All sweep arms share these values; only
    $c_{\mathrm{kl}}$ and $\lambda_{\mathrm{AFS}}$ differ across arms.}
    \label{tab:hyperparams}
    \small
    \begin{tabular}{p{0.27\linewidth}p{0.34\linewidth}p{0.30\linewidth}}
        \toprule
        \textbf{Component} & \textbf{Hyperparameter} & \textbf{Value} \\
        \midrule
        \multicolumn{3}{l}{\textit{World model: frozen backbone, trainable cross-attention + text-embedding adapter}} \\
        \midrule
        Architecture & WanModel, causal blockwise & 1.3B parameters \\
         & dim / num\_layers / num\_heads & 1536 / 30 / 12 \\
         & ffn\_dim / chunk\_size / num\_frames & 8960 / 3 / 21 \\
         & VAE compression $(T,H,W)$ & $(4,8,8)$ \\
        WM fine-tune & learning rate & $1{\times}10^{-5}$ \\
         & batch size & 1 \\
         & precision & bfloat16 \\
         & gradient clip & 1.0 \\
         & CFG dropout probability & 0.1 \\
         & PAT epochs per cycle & 7 \\
         & steps per cycle & $[500,1024]$ \\
         & PAT : passive replay ratio & 0.5 : 0.5 \\
         & cycle cadence & every 24 adversarial iterations \\
         & training hardware & $7\times$H200, DDP \\
        \midrule
        \multicolumn{3}{l}{\textit{PAT buffer: Prioritized Adversarial Trajectories}} \\
        \midrule
        Capacity & number of trajectories & $K=256$ \\
        Composite priority &
        \multicolumn{2}{p{0.64\linewidth}}{
        $z(\ell_{\mathrm{regret}}) + \lambda_{\mathrm{AFS}} z(\ell_{\mathrm{AFS}}) + \beta_{\mathrm{prog}} \Delta \ell_{\mathrm{regret}}$
        } \\
        $\lambda_{\mathrm{AFS}}$ & default value & 0.25 \\
        $\rho_{\mathrm{stale}}$ & staleness mixture weight & 0.1 \\
        Rescore frequency & after every WM update & --- \\
        \midrule
        \multicolumn{3}{l}{\textit{PPO adversarial policy: Phase 2, reference = frozen VPT-2x}} \\
        \midrule
        Reference policy & VPT-2x rl-from-early-game-2x & frozen \\
        Update cadence & PPO update frequency & every 16 episodes \\
        PPO epochs / minibatch & optimization setting & 4 / 256 \\
        Learning rate & policy optimizer & $3{\times}10^{-5}$ \\
        Clip $\epsilon$ & PPO clipping parameter & 0.2 \\
        Value coefficient & value loss weight & 0.5 \\
        Entropy coefficient & entropy bonus weight & 0.05 \\
        KL anchor $c_{\mathrm{kl}}$ & default, kl150 & 1.5 \\
        $\gamma$ / GAE $\lambda$ & return estimation & 0.99 / 0.95 \\
        Reward distribution & reward placement & terminal, one-shot \\
        Max grad norm & policy gradient clipping & 0.5 \\
        \midrule
        \multicolumn{3}{l}{\textit{Rollout / inference}} \\
        \midrule
        Seed chunks / horizon chunks & rollout setting & 2 / 2, i.e. 12 + 12 pixel frames \\
        $\max_t$ context window & latent context length & 21 latent frames \\
        History noise $\sigma$ & history corruption level & 0.05 \\
        CFG scale & classifier-free guidance & 1.5 \\
        Inference steps & sampler & 20, DPM \\
        Action window & action-conditioning window & 64 frames \\
        Image size & pixel resolution & $480 \times 832$ px \\
        \midrule
        \multicolumn{3}{l}{\textit{Sweep arms: six total, two-axis design}} \\
        \midrule
        KL anchor sweep &
        \multicolumn{2}{p{0.64\linewidth}}{
        $c_{\mathrm{kl}} \in \{0.5,1.0,1.5\}$ at $\lambda_{\mathrm{AFS}}=0.25$
        } \\
        $\lambda_{\mathrm{AFS}}$ sweep &
        \multicolumn{2}{p{0.64\linewidth}}{
        $\lambda_{\mathrm{AFS}} \in \{0.05,0.10\}$ at $c_{\mathrm{kl}}=1.0$
        } \\
        Holdout, static & evaluation rollouts & 32 frozen-VPT trajectories, seeds 2003--2034 \\
        Adversarial seeds & rollout seed rule & seed $=5000+\mathrm{iter}$ \\
        \bottomrule
    \end{tabular}
\end{table}

\section{Autoregressive Rollout Mechanics}
\label{app:rollout}

The window length $\mathrm{max}_t{=}21$ is structural: it is fixed by the Wan2.1 backbone's positional embeddings and attention shape (set during pretraining) and by the diffusion-forcing-aware loss the model was trained against. At inference, the window is filled with history latents followed by the $K{=}3$ target slots; any unused slots after the target are padded with pure Gaussian noise ($\sigma{=}1.0$), matching the ``unmasked future'' conditioning seen during training. Only the target slots are updated by the diffusion sampler at each denoising step; the padding remains inert and is never used as a prediction. After denoising, the predicted target chunk is appended to the history latents, the window slides forward by $K$ slots, and the next chunk is generated by repeating this procedure --- yielding chunk-level autoregressive rollout over $H$ predicted chunks.

\section{On the Asymmetric Form of the Trajectory Score}
\label{app:asymmetric_score}
The first two terms of Eq.~\ref{ref:eq_score} are z-normalized to make them dimensionless and balanced regardless of metric units, with $\lambda_{\mathrm{AFS}}$ controlling their relative weight. The third term, $\Delta\ell_{\mathrm{regret}}$, is left in raw latent-regret units. We adopt this asymmetric form for two reasons. First, $\Delta\ell_{\mathrm{regret}}$ is a per-trajectory temporal signal rather than a population statistic: z-normalizing it would require maintaining a separate $\Delta$-buffer of recent rescore deltas (additional state that does not change the rank ordering induced by the leading $z_{\mathcal{B}}$ term). Second, in pilot runs (200 iterations) the empirical standard deviation of $\Delta\ell_{\mathrm{regret}}$ across active buffer entries matched that of $\lambda_{\mathrm{AFS}}\,z_{\mathcal{B}}(\ell_{\mathrm{AFS}})$ within a factor of two; the three terms therefore contribute on similar empirical scales despite the formal asymmetry.

\section{Action Fingerprint Definition}
\label{app:fingerprint}
\paragraph{Tag assignment.}
A trajectory is labeled by concatenating an exclusive camera tier with any number of button tags:
\begin{itemize}[leftmargin=1.2em,topsep=0.1em,itemsep=0.05em]
    \item \emph{Camera tier} (at most one): \texttt{rot} if $\bar\theta(\tau) > 8^\circ$; else \texttt{look} if $\bar\theta(\tau) > 3^\circ$; else no camera tag.
    \item \emph{Button tags} (independent, multi-set): \texttt{fwd}, \texttt{sprint}, \texttt{atk}, \texttt{use}, \texttt{back}, \texttt{strafe} fire if the corresponding mean exceeds $0.3$; \texttt{jump} fires if its mean exceeds $0.2$ (jump presses are typically more bursty at $20$~FPS).
\end{itemize}
The label is then the concatenation of the camera tier and the active button tags in a fixed order, e.g.\ \texttt{rot+fwd+sprint+strafe}, or \texttt{still} if no tag fires.

\paragraph{Design rationale.}
Camera and buttons are kept on separate channels because they stress the world model in different ways (view-frustum motion vs.\ scene-state changes). The camera tier is exclusive because a single rollout window has one dominant camera regime; button tags are independent because Minecraft players combine button presses freely. The $0.3$/$0.2$ button thresholds correspond to sustained presses ($\geq 30\%$ / $\geq 20\%$ of the window) rather than single-frame taps, which we found to be the natural granularity for behavioral modes at our $24$-frame, $20$~FPS window. We deliberately use hard thresholds instead of clustering or a learned classifier: the rule has no tunable degrees of freedom, every tag has a literal definition, and the same constants are used across all buffers.

\section{Action Serialization for Cross-Attention Conditioning}
\label{app:action_serialization}

We serialize each chunk's per-frame MineRL action vector to a short text token (e.g., \texttt{forward jump look(up,left)}), run-length-encode consecutive identical tokens (e.g., \texttt{forward x3 | jump | idle}), and feed the resulting string through the frozen UMT5-XXL encoder. The cross-attention pathway then ingests these action embeddings as it would text embeddings in the original Wan2.1 text-to-video pretraining. Across both phases the VAE and UMT5-XXL encoder remain frozen; Phase~1 updates all $\sim$1.3B Wan2.1 parameters --- the spatial-temporal self-attention and MLP blocks, the cross-attention layers, and the linear adapter projecting UMT5 action-text embeddings into the diffusion-transformer's hidden dimension.

 \section{PROWL Training Loop}
\label{app:algorithm}

Algorithm~\ref{alg:prowl} formalizes the coordinator loop summarized in Section~\ref{sec:prowl_framework}: the alternating procedure by which the explorer policy $\pi_\phi$, the world model $\mathcal{W}_\theta$, and the PAT buffer $\mathcal{B}$ co-evolve over the course of Phase~2.

Each adversarial iteration follows a six-step cycle. \textbf{(1)~Explore:} the current explorer rolls out one episode in the environment, generating a candidate trajectory $\tau$. \textbf{(2)~Score:} the current world model predicts $\tau$ autoregressively, and three error signals are computed against the ground-truth rollout --- latent regret $R^{\mathrm{lat}}$, action-follow score $R^{\mathrm{afs}}$, and LPIPS --- which combine into the composite priority $s(\tau)$ (Eq.~\ref{ref:eq_score}). \textbf{(3)~Curate:} $\tau$ is inserted into $\mathcal{B}$, evicting the lowest-priority entry if $|\mathcal{B}|>K$, so the buffer holds the $K$ most informative discoveries seen so far. \textbf{(4)~Reward:} the latent regret is assigned as the terminal episode reward for PPO. \textbf{(5)~PPO update:} every $E$ episodes, the explorer is updated against this regret signal under a forward-KL anchor to the frozen reference policy $\pi_{\mathrm{ref}}$, with weight $c_{\mathrm{kl}}$ (Eq.~\ref{eq:ppo_loss}). \textbf{(6)~World-model cycle:} every $T_{\mathrm{wm}}$ iterations, the world model is fine-tuned on a 50/50 mixture of buffer trajectories (priority-weighted) and passive BASALT samples; afterwards, every trajectory in $\mathcal{B}$ is rescored under the updated model so that solved cases lose priority and unresolved ones rise.

This decoupling between high-frequency policy updates (every $E$ episodes) and lower-frequency world-model updates (every $T_{\mathrm{wm}}$ iterations) is what makes the loop tractable: the explorer searches against an approximately fixed model, while the model absorbs failures in batches and the buffer maintains a moving curriculum of unresolved weaknesses. The matched-compute Phase~2 (VPT-frozen) baseline corresponds to disabling step (5), pinning $\pi_\phi = \pi_{\mathrm{ref}}$ throughout while keeping the world-model update schedule identical.

\begin{algorithm}[h]
\caption{PROWL coordinator loop (one adversarial iteration).}
\label{alg:prowl}
\begin{algorithmic}[1]
\Require world model $\mathcal{W}_\theta$ (Phase-1 init), explorer $\pi_\phi$ (init = $\pi_{\mathrm{ref}}$),
PAT buffer $\mathcal{B}$, sweep weights $c_{kl}$, $\lambda_{\mathrm{AFS}}$
\Repeat
  \State \textbf{(1) Explore.}\;$\tau \gets$ rollout $\pi_\phi$ in environment for one episode
  \State \textbf{(2) Score.} \;Run $\mathcal{W}_\theta$ on $\tau$; compute $R^{\mathrm{lat}}(\tau), R^{\mathrm{afs}}(\tau)$
  \State \hspace{1.7em}composite priority $s(\tau) \gets z(R^{\mathrm{lat}}) + \lambda_{\mathrm{AFS}}\,z(R^{\mathrm{afs}}) + \beta\,\Delta R^{\mathrm{lat}}$
  \State \textbf{(3) Curate.}\;Insert $\tau$ into $\mathcal{B}$; evict lowest-priority entry if $|\mathcal{B}|>K$
  \State \textbf{(4) Reward.}\;Episode reward $r(\tau) = R^{\mathrm{lat}}(\tau)$ (terminal); store $(\tau,r)$ for PPO
  \State \textbf{(5) PPO update} (every $E$ episodes):
  \State \hspace{1.7em} $\phi \gets \phi + \eta_\pi\,\nabla_\phi\bigl[\,\mathbb{E}\,[\hat A] \;-\; c_{kl}\,\mathrm{KL}(\pi_{\mathrm{ref}} \Vert \pi_\phi)\,\bigr]$
  \State \textbf{(6) WM cycle} (every $T_{\mathrm{wm}}$ iterations):
  \State \hspace{1.7em}sample $\mathcal{D} \;=\;$ 50\% from $\mathcal{B}$ (priority-weighted) $\;+\;$ 50\% passive replay
  \State \hspace{1.7em}fine-tune $\mathcal{W}_\theta$ on $\mathcal{D}$ for $S$ steps (cross-attention only)
  \State \hspace{1.7em}\textbf{rescore} every $\tau\in\mathcal{B}$ under the new $\mathcal{W}_\theta$ and refresh priorities
\Until{iteration budget exhausted}
\State \Return $\mathcal{W}_\theta$
\end{algorithmic}
\end{algorithm}
\vspace{-1.5em}

\section{AFS-EPE vs.~Appearance Metrics: When Motion Fidelity Hides}
\label{app:afs_justification}

A natural question is whether the optical-flow term in our trajectory score (Eq.~\ref{ref:eq_score}) is necessary, given that predicted frames can be compared to ground truth using standard appearance-based metrics --- latent regret in VAE space, perceptual similarity via LPIPS, or structural similarity via SSIM. This appendix shows empirically that on action-heavy held-out frames, all three appearance metrics can remain tied or near-tied between checkpoints while AFS-EPE shows a $2$--$3\times$ gap that corresponds visually to incorrect motion. The flow term is therefore not redundant: it captures a class of failures invisible to appearance.

\begin{figure}[!h]
\centering
\includegraphics[width=\linewidth]{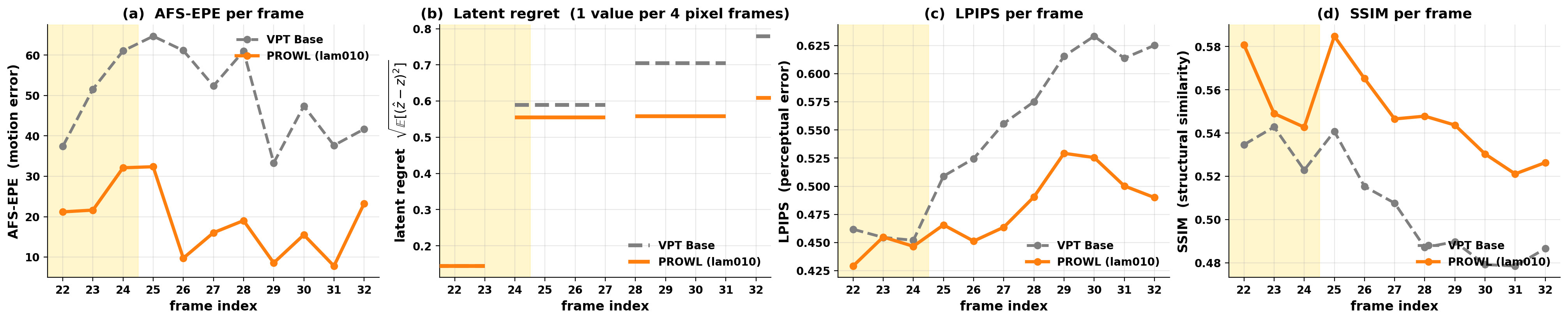}
\caption{\textbf{AFS-EPE catches motion failures invisible to appearance metrics.} Same trajectory as Figure~\ref{fig:afs_filmstrip}. Yellow band: action-heavy frames (forward+left + camera rotation, frames~22--24). \textbf{(a)}~Per-frame AFS-EPE shows a $2$--$3\times$ gap between VPT Base and PROWL throughout the action-heavy frames. \textbf{(b)}~Latent regret (one value per VAE-latent frame, hence step-shaped) is nearly tied during the action-heavy frames and only diverges later. \textbf{(c)}~LPIPS shows the two checkpoints nearly tied through frame~24, with the gap appearing only as early motion errors compound. \textbf{(d)}~SSIM shows the two checkpoints nearly tied (and even VPT Base briefly higher at frames~23--24) during action-heavy frames despite the large AFS-EPE gap.}
\label{fig:afs_metrics}
\end{figure}

\paragraph{Setup.} We evaluate one held-out trajectory from the VPT reference policy containing a heavy combined-action sequence: forward+left locomotion combined with sustained camera rotation (cam$(-40,-21)\to$ cam$(-22,-15)$) over frames~22--24, marked as the yellow band in Figure~\ref{fig:afs_metrics}. This kind of compound motion is where action-conditioning is most stressed: the model must produce the correct view-frustum motion on top of the correct scene-relative motion. Frames~25--32 release into smaller camera-only motion. We compare the Phase~1 VPT Base checkpoint against PROWL~(\texttt{lam010}) under identical actions.

\begin{figure}[!h]
\centering
\includegraphics[width=0.8\linewidth]{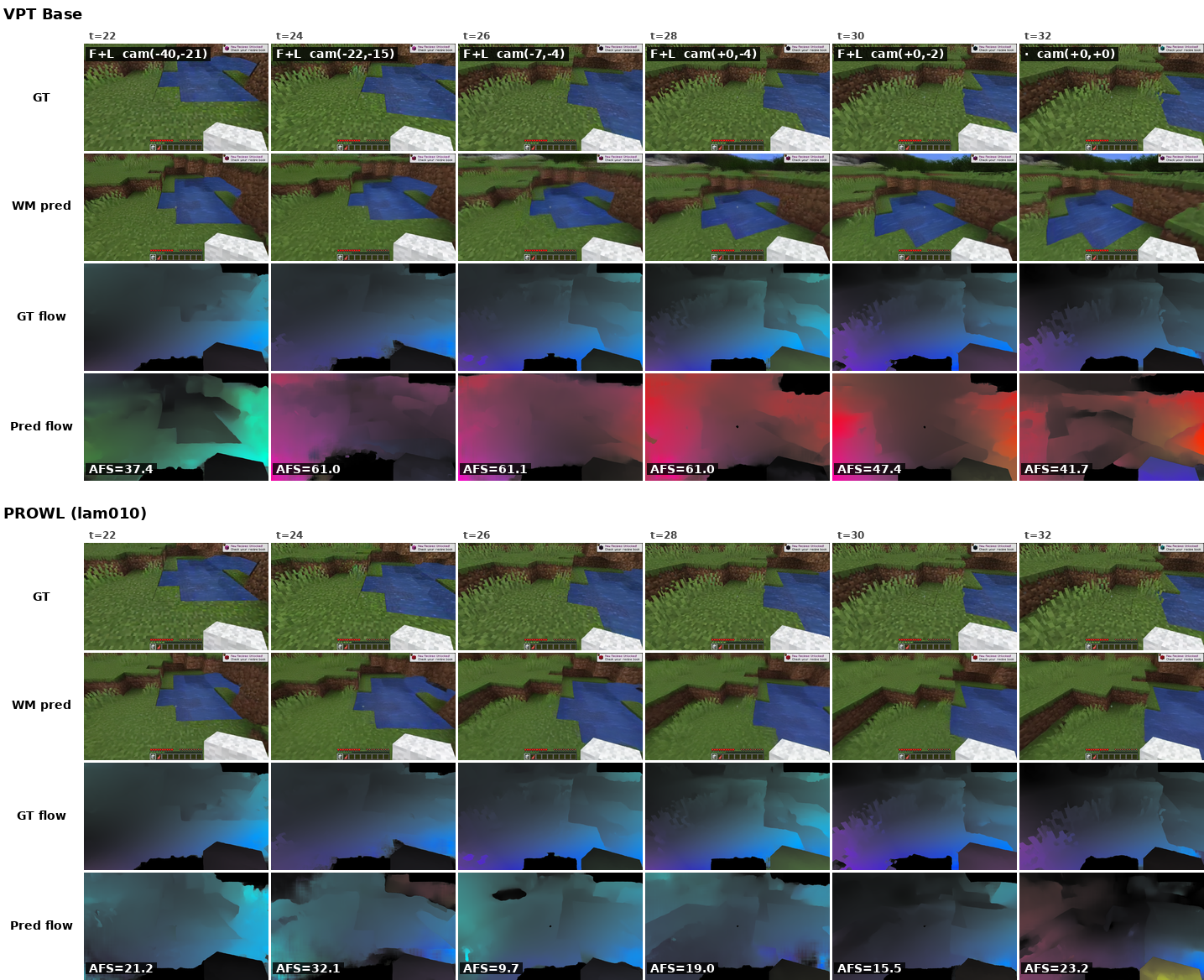}
\caption{\textbf{Qualitative comparison: VPT Base vs.~PROWL under identical actions.} Frames~22--32 of a held-out VPT-reference trajectory containing forward+left + camera rotation through frames~22--24 (action-heavy region). Each group shows ground-truth frames, predicted frames, ground-truth optical flow, and predicted flow. VPT Base's predicted flow during the action-heavy frames is incoherent (red/magenta blobs); PROWL's tracks the GT flow direction (blue/teal gradients). Per-frame AFS values are inset in the predicted-flow row.}
\label{fig:afs_filmstrip}
\end{figure}

\paragraph{Result.} Across all three appearance metrics (Figure~\ref{fig:afs_metrics}b--d), the two checkpoints are nearly tied during the heaviest action frames~22--24, with divergence appearing only afterward. AFS-EPE (Figure~\ref{fig:afs_metrics}a), in contrast, shows a sharp and immediate gap during exactly the action-heavy frames: VPT Base reaches motion errors of $60+$ pixels per frame while PROWL stays around $20$--$30$. The qualitative filmstrip in Figure~\ref{fig:afs_filmstrip} confirms this visually: VPT Base's predicted optical flow during frames~22--24 is incoherent (red/magenta blobs, AFS=37--61), while PROWL's predicted flow tracks the ground-truth flow direction (blue/teal gradients, AFS=21--32).

\paragraph{Why this happens.} Appearance metrics aggregate over the full frame and reward visually plausible reconstructions even when the \emph{direction} of motion is wrong. A predicted frame that contains the right textures and roughly correct geometry will score well on LPIPS and SSIM even if it represents the result of a wrong action --- a model that ``freezes'' or ``lags'' under heavy motion can still produce visually plausible static-looking content. Latent regret has the same blind spot in latent space: the latent can lie in a plausible region without corresponding to the correct motion. AFS-EPE penalizes this directly by computing per-pixel optical-flow disagreement, which is sensitive to both magnitude and direction of motion.

\paragraph{Implication for the trajectory score.} The two error terms in Eq.~\ref{ref:eq_score} are complementary by design: $\ell_{\mathrm{regret}}$ catches global latent-plausibility errors while $\ell_{\mathrm{AFS}}$ catches motion-fidelity errors that all three appearance metrics miss. Removing $\ell_{\mathrm{AFS}}$ would leave the adversary blind to a substantial class of failures during exactly the action-heavy regimes that PROWL is designed to expose --- and hence blind to the failure mode that distinguishes a model that follows actions from one that merely produces plausible-looking video.

\end{document}